\documentclass[journal]{IEEEtran}
\usepackage{booktabs} 
\usepackage{graphicx}
\usepackage{amsfonts}
\usepackage{dsfont}
\usepackage{graphicx, epsfig,amsmath,amssymb,latexsym,graphics,float}
\usepackage{setspace,subfig}
\usepackage{citesort}
\usepackage{float}
\usepackage[ruled,vlined,linesnumbered]{algorithm2e}
\usepackage{booktabs}
\usepackage{lipsum}
\usepackage{multicol}
\usepackage{url}
\usepackage{multirow}
\usepackage[ruled]{algorithm2e} 

\begin{document}
\title{A Particle Filter based Multi-Objective Optimization Algorithm: PFOPS}
\author{Bin~Liu$^{\star}$, Yaochu~Jin
\thanks{$^{\star}$Corresponding author. B. Liu is with School of Computer Science, Nanjing University of Posts and Telecommunications, and Jiangsu Key Lab of Big Data Security $\&$ Intelligent Processing, Nanjing,
Jiangsu, 210023 China.
E-mail: bins@ieee.org. Yaochu Jin is with Department of Computer Science, University of Surrey, Guildford, Surrey, GU2 7XH, UK. Email: yaochu.jin@surrey.ac.uk.
}
\thanks{Manuscript submitted to Arxiv at Aug. 28, 2018; revised at Oct. 28, 2018.}}

%
%

\markboth{Journal Name, ~Vol.~XX, No.~X, XX~2018}%
{Liu, et.al: PFPS Algorithm for Multi-Objective Optimization}
%


\maketitle

\begin{abstract}
This paper is concerned with a recently developed paradigm for population-based optimization, termed particle filter optimization (PFO). This paradigm is attractive in terms of coherence in theory and easiness in mathematical analysis and interpretation. Current PFO algorithms only work for single-objective optimization cases, while many real-life problems involve multiple objectives to be optimized simultaneously. To this end, we make an effort to extend the scope of application of the PFO paradigm to multi-objective optimization (MOO) cases. An idea called path sampling is adopted within the PFO scheme to balance the different objectives to be optimized. The resulting algorithm is thus termed PFO with Path Sampling (PFOPS). The validity of the presented algorithm is assessed based on three benchmark MOO experiments, in which the shapes of the Pareto fronts are convex, concave and discontinuous, respectively.
\end{abstract}
\begin{IEEEkeywords}
particle filtering, path sampling, multi-objective optimization, derivative-free optimization, Pareto front.
\end{IEEEkeywords}
\IEEEpeerreviewmaketitle
\section{Introduction}
This paper presents a novel MOO algorithm based on the particle filter (PF). The PFO methods belong to the class of population-based derivative-free optimization methods \cite{zhou2008particle,stinis2012stochastic,liu2017posterior,liu2016particle,zhou2013sequential}. Different from the meta-heuristics based evolutionary computation (EC) methods \cite{kennedy2011particle,mitchell1998introduction,dorigo2005ant}, the PFO paradigm is developed based on Bayesian statistics instead of meta-heuristics. As a class of Sequential Monte Carlo (SMC) methods, the convergence of PF under mild conditions has been proved \cite{chopin2004central,chopin2004central}. This result also holds for the PFO methods. In contrast, the class of EC methods is weak in theory due to the lack of a strict and coherent mathematical foundation.

Current PFO methods only work for single-objective optimization (SOO) problems,
while many real-life problems involve multiple objectives to be optimized simultaneously.
To this end, we make an effort to extend the scope of application of PFO to multi-objective
cases. The key insight adopted here is that, if we can construct a series of target distributions that can balance the multiple objectives and make the degree of this balance controllable, then by simulating these distributions via SMC, we can evaluate the Pareto optimal solutions based on the samples yielded from simulations of these target distributions.
We borrow an idea called path sampling to construct the target distributions. We show that the resulting method can handle multiple objectives in an elegant and easy-to-implement way while maintaining the theoretical soundness of the PFO framework. Note that the idea of path sampling is originally developed for estimating the marginal likelihoods of candidate models in the context of Bayesian model comparison \cite{gelman1998simulating}.

The remainder of this paper is organized as follows. Section \ref{sec:moo} briefly introduces the necessary background on MOO problems. Section \ref{sec:alg} describes the proposed algorithm and discusses the relationships between it and the other related works. Section \ref{sec:simu} presents the simulation results, and finally, Section \ref{sec:conclusion} concludes.
\section{Background on MOO}\label{sec:moo}
Let us consider an optimization problem with $M$ objectives as follows
\begin{equation}\label{eqn:moo}
\min F(\textbf{\mbox{x}})=(f_1(\textbf{\mbox{x}}),f_2(\textbf{\mbox{x}}),\ldots,f_M(\textbf{\mbox{x}})),\ \mbox{subject to}\ \textbf{\mbox{x}}\in X,
\end{equation}
where $f_i: X\rightarrow\mathbb{R}$ denotes the $i$th real-valued continuous objective function to be minimized, $\textbf{\mbox{x}}=(x_1,\ldots,x_d)$ is a $d$ dimensional decision vector with value space $X$.
The difficulty in resolving (\ref{eqn:moo}) results from the conflicts among the objectives $f_1, \ldots, f_M$, which means that a decision that decreases the value of $f_m, m\in\{1,\ldots,M\}$ may increase that of $f_n, n\neq m, n\in\{1,\ldots,M\}$. As a consequence, there is no single decision that minimizes all the objectives simultaneously. A basic idea to deal with this conflict is to find a set of optimal decisions that trade-off among these different objectives, which motivates the concept of Pareto dominance. Given two decisions $\textbf{\mbox{x}},\textbf{\mbox{x}}^{\prime}\in X$, we say $\textbf{\mbox{x}}$ (Pareto) dominates $\textbf{\mbox{x}}^{\prime}$, denoted by $\textbf{\mbox{x}}\prec \textbf{\mbox{x}}^{\prime}$, iff $f_m(\textbf{\mbox{x}})\leq f_m(\textbf{\mbox{x}}^{\prime})$, $\forall m\in\{1,\ldots,M\}$ and $\exists m\in\{1,\ldots,M\}$, $f_m(\textbf{\mbox{x}})< f_m(\textbf{\mbox{x}}^{\prime})$. A decision $\textbf{\mbox{x}}^{\star}\in X$ is called (globally) \emph{Pareto optimal} if there is no $\textbf{\mbox{x}}\in X$ such that $\textbf{\mbox{x}}\prec \textbf{\mbox{x}}^{\star}$. The set of all the Pareto optimal decisions is called the Pareto set ($PS$). The set of all Pareto optimal objective vectors, $PF=\{\textbf{\mbox{y}}\in\mathbb{R}^m|\textbf{\mbox{y}}=F(\textbf{\mbox{x}}),\textbf{\mbox{x}}\in PS\}$, is called the Pareto front \cite{deb2014multi}. Most existent solutions for such MOO problems are based on EC methods, see details in \cite{zhang2007moea,deb2002fast,marler2004survey}. Here we propose an alternative algorithm based on Bayesian statistics in what follows.
\section{PFOPS: The Proposed Algorithm}\label{sec:alg}
Here we adopt a Bayesian probabilistic viewpoint to investigate the optimization problem. According to this viewpoint, our belief on the minimizer $\textbf{\mbox{x}}^{\star}$ of an objective function $f$ can be quantitatively measured by a probability density function (pdf), say $\tilde{\pi}\varpropto\exp(-f_m)$. If we can simulate $\tilde{\pi}$ by continuously drawing random samples from it, then $\textbf{\mbox{x}}^{\star}$ can be evaluated based on the samples yielded from that simulation. We can improve this viewpoint by borrowing the idea of simulated annealing (SA) via designing a series of target pdfs that asymptotically converge to the set of global optima. Then, through simulating the target pdfs by e.g., Markov Chain Monte Carlo (MCMC) as in SA \cite{kirkpatrick1983optimization} or SMC as in \cite{liu2017posterior,zhou2013sequential}, one can get a set of random samples whose empirical distribution also converge asymptotically to the set of global optima. Thus the global optima can be evaluated based on the yielded samples. The above discussion is only limited to SOO problems. Now we extend it to the context of MOO.
\subsection{The details of the algorithm}
As mentioned in Section \ref{sec:moo}, the problem of MOO can be formulated as a task of searching a set of Pareto optimal decisions, each of which corresponds to a certain degree of tradeoff among the objectives to be optimized. The basic insight adopted here is that, if we can build up a series of proxy target pdfs, each corresponding to a specific amount of balance among the objectives to be optimized, then, by simulating these proxy pdfs one by one, we can estimate the Pareto optimal decisions along with the corresponding Pareto front based on the simulated samples. We utilize the idea of path sampling to construct the proxy target distributions, then adopt PF to simulate them. In the present work, we focus on 2-objective problems to avoid complications in illustrating the proposed concepts. Given objective functions $f_1(\textbf{\mbox{x}})$ and $f_2(\textbf{\mbox{x}})$, we borrow the idea of path sampling \cite{gelman1998simulating} to construct a series of target pdfs $\tilde{\pi}_1(\textbf{\mbox{x}}),\tilde{\pi_2}(\textbf{\mbox{x}}),\ldots,\tilde{\pi}_K(\textbf{\mbox{x}})$ as follows
\begin{eqnarray}\label{eqn:target_pdfs}
\tilde{\pi}_k(\textbf{\mbox{x}})&\triangleq&\frac{\pi_k(\textbf{\mbox{x}})}{C_k},\ k=1,2,\ldots,K, \\
\pi_k(\textbf{\mbox{x}})&=&\exp\{-[(1-\lambda_k)f_1(\textbf{\mbox{x}})+\lambda_kf_2(\textbf{\mbox{x}})]\},
\end{eqnarray}
where $K$ is the number of target pdfs, $0=\lambda_1<\lambda_2<,\ldots,<\lambda_K=1$, $C_k$ is a normalizing constant which ensures $\tilde{\pi}_k(\textbf{\mbox{x}})$ to be a qualified pdf whose integral equals 1. According to Eqn.(\ref{eqn:target_pdfs}), $\tilde{\pi}_1(\textbf{\mbox{x}})$ and $\tilde{\pi}_K(\textbf{\mbox{x}})$ are completely dependant on $f_1(\textbf{\mbox{x}})$ and $f_2(\textbf{\mbox{x}})$, respectively, and  $\tilde{\pi}_2(\textbf{\mbox{x}}),\ldots,\tilde{\pi}_{K-1}(\textbf{\mbox{x}})$ are intermediate pdfs that connects $\tilde{\pi}_1(\textbf{\mbox{x}})$ with $\tilde{\pi}_K(\textbf{\mbox{x}})$. Each pdf corresponds to a specific degree of balance, parameterized by $\lambda_k$, between $f_1(\textbf{\mbox{x}})$ and $f_2(\textbf{\mbox{x}})$.
\begin{algorithm}[!htb]
\caption{\label{algo:MOO}The proposed MOO algorithm: PFOPS. In this table, $\textbf{\mbox{x}}_k^{\star}$ denotes the $k$th Pareto optimal decision generated, ``Unif" uniform distribution, $N$ the sample size of PF,  $\mathcal{N}(\cdot|x,\Sigma)$ normal distribution with mean $x$ and variance $\Sigma$, $\delta(\cdot)$ the delta-mass function located at 0, ``$\sim$" means ``distributed as", $\forall i$ is an abbreviation of $\forall i\in\{1,\ldots,N\}$.}
Initialization: Define the target pdfs according to Eqn.(\ref{eqn:target_pdfs}); Draw i.i.d. samples
    $\textbf{\mbox{x}}^i\thicksim\mbox{Unif}(X)$,
    for $\forall i$. Initialize $\hat{PS}$ as an empty set\;
\For{$k=1,\ldots, K$ }{
Find $j=\underset{i\in\{1,\ldots,N\}}{\max}\pi_k(\textbf{\mbox{x}}^i)$ and set $\textbf{\mbox{x}}_k^{\star}=\textbf{\mbox{x}}_j$\;
Importance weighting: for $\forall i$, set
    \begin{displaymath}
\hat{\omega}_{k}^i=\left\{\begin{array}{ll}
\pi_k(\textbf{\mbox{x}}^i), \ \mbox{if}\quad k=1;\\
\pi_k(\textbf{\mbox{x}}^i)/\pi_{k-1}(\textbf{\mbox{x}}^i), \ \mbox{otherwise}. \end{array} \right.
\end{displaymath}
Normalize the importance weights: for $\forall i$, set
    $\omega_{k}^i=\hat{\omega}_{k}^i/\sum_{i=1}^N\hat{\omega}_{k}^i$\;
Resampling: generate $N$ i.i.d. samples $\{\tilde{\textbf{\mbox{x}}}^i\}_{i=1}^N$ by setting $\tilde{\textbf{\mbox{x}}}^i=\textbf{\mbox{x}}^j$ with probability $\omega_k^j$, $j=1,\ldots,N$. Then set $\textbf{\mbox{x}}^i=\tilde{\textbf{\mbox{x}}}^i$, $\omega_k^i=1/N$, for $\forall i$\;
Componentwise Metropolis sampling\;
\For{$i=1,\ldots, N$ }{
 \For{$j=1,\ldots, d$}{
   Set $\textbf{\mbox{x}}^{\prime}=\textbf{\mbox{x}}^i$, where $\textbf{\mbox{x}}^{\prime}\triangleq(x_1^{\prime},\ldots,x_d^{\prime})$\;
   Modify the $j$th dimension of $\textbf{\mbox{x}}^{\prime}$ by setting $x_j^{\prime}\sim \mathcal{N}(\cdot|x_j^i,\Sigma)$. If $\pi_k(\textbf{\mbox{x}}^{\prime})>\pi_k(\textbf{\mbox{x}}_k^{\star})$, set $\textbf{\mbox{x}}_k^{\star}=\textbf{\mbox{x}}^{\prime}$\;
 Calculate acceptance probability
\begin{displaymath}
\rho=\min\left\{\pi_k(\textbf{\mbox{x}}^{\prime})/\pi_k(\textbf{\mbox{x}}^i),1\right\}.
\end{displaymath}
Accept
\begin{displaymath}
\textbf{\mbox{x}}^i=\left\{\begin{array}{ll}
\textbf{\mbox{x}}^{\prime}, \mbox{with probability}\;\rho;\\
\textbf{\mbox{x}}^i, \mbox{with probability}\;1-\rho. \end{array} \right.
\end{displaymath}
}
}
Record $\textbf{\mbox{x}}_k^{\star}$ into $\hat{PS}$\;
}
Set $\hat{PF}=\{\textbf{\mbox{y}}\in\mathbb{R}^m|\textbf{\mbox{y}}=F(\textbf{\mbox{x}}),\textbf{\mbox{x}}\in \hat{PS}\}$\;
Remove any $\textbf{\mbox{y}}\in\hat{PS}$ (if it exists) that satisfies $\exists \textbf{\mbox{x}}\in\hat{PS}, \textbf{\mbox{x}}\prec \textbf{\mbox{y}}$ from $\hat{PS}$. Then update $\hat{PF}$ correspondingly\;
Output $\hat{PS}$ and $\hat{PF}$ as the estimated Pareto set and Pareto front, respectively.
\end{algorithm}

Note that the choice of target pdfs defined by Eqn. (2)-(3) is not restrictive and does not characterize the presented methodology, as other forms of pdfs can be used instead if they fit the problem better. For example, in the experiment presented in Section \ref{sec:concave}, a different form of $\pi_k(\textbf{\mbox{x}})$, see Eqn.(\ref{eqn:Tchbycheff}), is used instead to take into account the Pareto front being concave or discontinuous.

Given the target pdfs, a PF based sampling procedure is used to simulate these pdfs. Based on the yielded simulated samples, we estimate the Pareto optimal decisions as well as the corresponding Pareto front. Due to the space limitation, we bypass the introduction of any background of PF, while referring readers to \cite{arulampalam2002tutorial,doucet2000sequential,liu1998sequential} for details. A summarization of the proposed PFOPS method is presented in Algorithm \ref{algo:MOO}.

In Algorithm \ref{algo:MOO}, the importance weighting, normalization of importance weights and resampling are just typical operations of PF \cite{arulampalam2002tutorial}. Through the resampling step, we eliminate/duplicate samples with low/high importance weights respectively. This operation is effective for avoiding the issue of particle degeneracy. A metropolis sampling step is added for strengthening particle diversity. We select to perform componentwise metropolis sampling since this can make the resulting algorithm scalable to high-dimensional problems \cite{haario2005componentwise}. As the target distribution of the metropolis sampling step is the same as that of the importance sampling, although new samples will be generated, the distribution of the resulting samples keeps invariant. $\Sigma$ is a scalar parameter, whose value is preset to be 1 in our experiments.
\subsection{Related work}
The probabilistic nature of our method makes it closely related with the class of estimation of distribution algorithms (EDAs) \cite{liu2018improving,shim2012hybrid,karshenas2014multiobjective}.
They both use probabilistic models to lead the search towards a more promising area in the decision space. The parametric model used in EDAs needs to be fitted and updated in each iteration, while each model fitting operation brings an additional optimization task to be resolved. By contrast, the proposed algorithm here totally frees the user from the issue of repeated model fitting, since the probabilistic models, i.e., the target pdfs here, are precisely defined beforehand and thus do not need to be fit.

The present work also has connections to a class of algorithms called Bayesian optimization (BO) \cite{shahriari2016taking,hernandez2016predictive}, since they are both Bayesian methods used for dealing with optimization problems. The underpinning assumption in these BO methods is that the objective function is too expensive to be evaluated. A basic operation to reduce the number of expensive function evaluations, which also characterizes the BO methods, is to repeatedly fit a model such as the Gaussian process to approximate the objective function based on a growing number of real function evaluations that have been done. By contrast, the proposed PFOPS algorithm is a kind of population-based methods that do not take an assumption of expensive objective function evaluations. Meanwhile, it has no operations of model fitting included.

Finally, the proposed algorithm is surely related with the other existent PFO methods in e.g. \cite{zhou2008particle,stinis2012stochastic,liu2017posterior,liu2016particle,zhou2013sequential}. In short, the present work can be seen as an extension of existing PFO methods targeting the MOO problem.
\section{Performance Evaluation}\label{sec:simu}
We assessed the behavior of the proposed PFOPS algorithm via numerical experiments. We used one celebrated EC based MOO algorithm, NSGA-II \cite{deb2002fast}, as a benchmark of performance comparison. We considered three two-objective optimization problems, in which the shapes of the Pareto fronts are convex, concave and
discontinuous, respectively.
\subsection{Convex Pareto Front Case}
In this case, we set $\textbf{\mbox{x}}=(x_1,x_2)$, $f_1(\textbf{\mbox{x}})\triangleq x_1^2+x_2^2$, $f_2(\textbf{\mbox{x}})\triangleq (x_1-5)^2+(x_2-5)^2$, $x_1\in[-5,10]$ and $x_2\in[-5,10]$. For both PFOPS and NSGA-II, we use two parameter settings, called sufficient-sampling and undersampling here, to initialize them. In the sufficient-sampling setting, we allocate for the algorithm enough iterations to run and a much bigger population size for use, in order to check its performance limit. In contrast, the undersampling setting means a limited number of iterations and a far smaller population size allocated. By doing so, we hope we can generalize results observed here to high dimensional cases for which the undersampling setting is the only choice.

Specifically, we set $K=N=100$ in the oversampling setting of PFOPS; and set $K=20$ and $N=5$ in the undersampling setting. In both settings, the value of $\lambda_k$ in Eqn.(3) is equal-interval sampled between 0 and 1, and the optional step is not executed. For NSGA-II, we use $\sharp$pop and $\sharp$Gen to denote its population size and the number of generations, respectively. In the oversampling setting $\sharp$pop=100 and $\sharp$Gen=100, while for the undersampling case, $\sharp$pop=20 and $\sharp$Gen=5. Now we let $\sharp$fe denote the number of fitness evaluations, then we have $\sharp$fe=$2\times K\times N$ for PFOPS and $\sharp$fe=2$\times\sharp$pop$\times$$\sharp$Gen for NSGA-II. Then we can see that, in both settings, the computational burden for each algorithm (in terms of $\sharp$fe) is maintained at the same level. A direct wall clock running time comparison is also listed in Table \ref{Table:time}, which re-confirms that a fair comparison is made here.
\begin{table*}[!phtb]\caption{Wall clock running time averaged over 10 independent runs. The unit of time is seconds.}\label{Table:time}
\centering
\begin{tabular}{ccc}
\hline  & sufficient-sampling case & undersampling case \\
\hline PFOPS & 5.4951 & 0.1158 \\
\hline NSGA-II & 5.2153 & 0.4171\\
\hline
\end{tabular}
\end{table*}
\begin{figure}[!htb]
\begin{tabular}{c}
\centerline{\includegraphics[width=3.55in,height=2.6in]{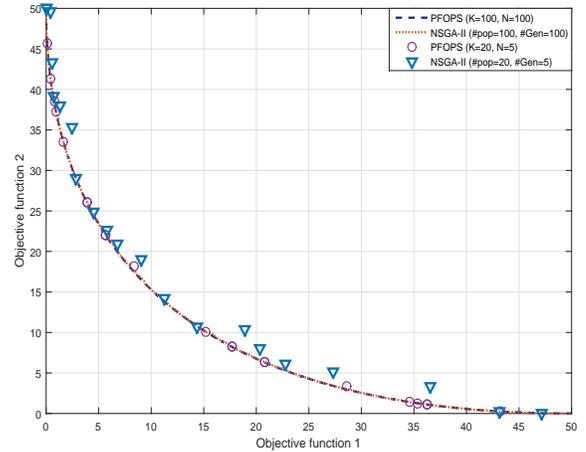}}
\end{tabular}
\caption{The Pareto fronts yielded by PFOPS and NSGA-II. $\sharp$pop and $\sharp$Gen are hyperparameters of NSGA-II representing the population size and the number of generations, respectively.}\label{fig:pareto_front}
\end{figure}

We do experiments by running each algorithm under both of the sufficient-sampling and undersampling settings. An illustration of the estimated Pareto fronts obtained from a typical experiment is presented in Fig.\ref{fig:pareto_front}. It shows that, under the sufficient-sampling setting, the Pareto fronts of PFOPS and NSGA-II almost coincide with each other. It indicates that PFOPS will perform as well as NSGA-II under the sufficient-sampling setting if the Pareto front is convex. Fig.\ref{fig:pareto_front} also shows that, under the undersampling setting, PFOPS performs better than NSGA-II, as the deviation between samples given by PFOPS and the true Pareto front is significantly smaller than that of NSGA-II. Note that here the so-called true Pareto front is exactly an estimate of it given by PFOPS and NSGA-II under the sufficient-sampling setting.
\subsection{Concave and Discontinuous Pareto Front Cases}\label{sec:concave}
The aim of the experiment presented here is to test whether the proposed PFOPS algorithm is capable of handling MOO problems that own a non-convex or a discontinuous Pareto front. For such problems, the weighted-sum type trade-off between the objectives, as shown in Eqn.(3), does not work, as its associated function curve is convex and continuous. Hence, an alternative design of $\pi_k(\textbf{\mbox{x}})$ is employed here, which is shown as follows
\begin{equation}\label{eqn:Tchbycheff}
\pi_k(\textbf{\mbox{x}})=\exp(-\max\{(1-\lambda_k)|f_1(\textbf{\mbox{x}})-z_1^{\star}|,\lambda_k|f_2(\textbf{\mbox{x}})-z_2^{\star}|\}),
\end{equation}
where $z_1<\min f_1, z_2<\min f_2$, and $\textbf{\mbox{z}}^{\star}=(z_1^{\star},z_2^{\star})$ is termed Utopian point. We consider here two benchmark test functions for MOO, i.e., a 2-dimensional Fonseca-Fleming function \cite{fonseca1995overview} defined as follows
\begin{eqnarray*}
f_1(\textbf{\mbox{x}})&=&1-\exp[-\sum_{i=1}^2(x_i-1/\sqrt{2})], \\ f_2(\textbf{\mbox{x}})&=&1-\exp[-\sum_{i=1}^2(x_i+1/\sqrt{2})],
 \end{eqnarray*}
 where $-4\leq x_i\leq 4, i=1,2$, and the Kursawe function \cite{kursawe1990variant} defined as follows
 \begin{eqnarray*}
f_1(\textbf{\mbox{x}})&=&\sum_{i=1}^2[-10\exp(-0.2\sqrt{x_i^2+x_{i+1}^2})] \\ f_2(\textbf{\mbox{x}})&=&\sum_{i=1}^3\left[|x_i|^{0.8}+5\sin(x_i^3)\right],
 \end{eqnarray*}
where $-5\leq x_i\leq 5, 1\leq i\leq 3$.
\begin{figure}[!htb]
\begin{tabular}{c}
\centerline{\includegraphics[width=3.55in,height=2.6in]{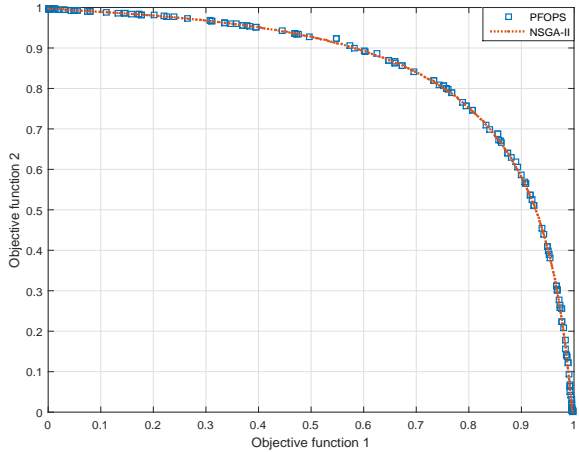}}
\end{tabular}
\caption{The estimated Pareto fronts by PFOPS and NSGA-II for an MOO problem which uses the 2D Fonseca-Fleming function as the test function. This case is characterized by a concave Pareto front as shown in the Figure.}\label{fig:concave}
\end{figure}
\begin{figure}[!htb]
\begin{tabular}{c}
\centerline{\includegraphics[width=3.55in,height=2.6in]{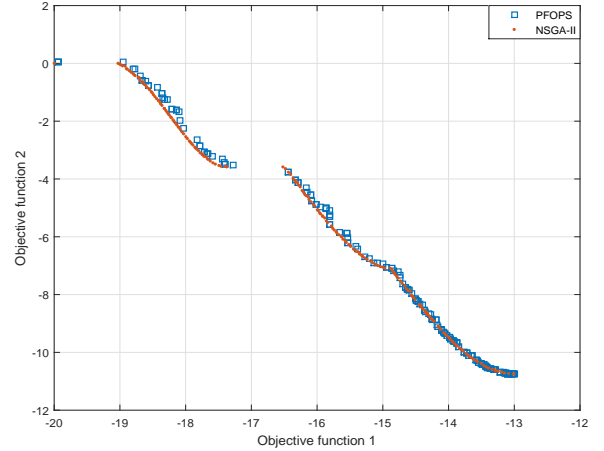}}
\end{tabular}
\caption{The estimated Pareto fronts by PFOPS and NSGA-II for an MOO problem which uses the Kursawe function as the test function. This case is characterized by a discontinuous Pareto front as shown in the Figure.}\label{fig:Kursawe}
\end{figure}

In the experiment, we set for the former case $z_1^{\star}=z_2^{\star}=-1$, and for the latter $z_1^{\star}=-21, z_2^{\star}=-13$. For both cases, we set $K=200, N=500$ for initializing the PFOPS algorithm; and $\sharp$pop=200, $\sharp$Gen=500 for NSGA-II. The experimental result for these two cases are shown in Figs.\ref{fig:concave} and \ref{fig:Kursawe}, respectively. As is shown, for both cases, the PFOPS gives an estimate of the Pareto front that is comparable to the NSGA-II algorithm.
\section{Concluding Remarks}\label{sec:conclusion}
In this paper, we extended the scope of application of the PFO paradigm to MOO cases by proposing a novel PF based MOO algorithm called PFOPS. This algorithm discriminates itself with the other related works by adopting a path sampling based mechanism to balance the differing objectives within the PFO paradigm. Experimental results show that it performs better than NSGA-II for the considered convex Pareto front case based on the undersampling setting and its performance is comparable to NSGA-II in the other two cases, in which the shape of the Pareto front is concave and discontinuous, respectively.

The present work represents an initial attempt to formulate the MOO problem as a state filtering task and then to handle it using state filtering methods like PF here. This filtering perspective has already been explored in developing PF based SOO algorithms, such as in \cite{zhou2008particle,stinis2012stochastic,liu2017posterior,liu2016particle,zhou2013sequential}. Combining these previous studies with the present work here, we can see a big picture, in which a unified PFO framework has been built up, which can handle both SOO and MOO problems in a consistent way. The basic operation included in this framework consists of two parts. The first is to construct a series of target pdfs and the second is to sample from the constructed pdfs. This framework is different from the meta-heuristics based EC methods. Compared with EC methods, this framework is theoretically attractive thanks to its probabilistic nature that makes it own a stronger tie to mathematics especially statistics.

Finally, we point out two interesting future works following this line of research. One is to explore efficient mechanisms to construct the target pdfs for problems with more than two objectives and the other is to improve the present algorithm by adopting the simulated annealing strategy.
\bibliographystyle{IEEEbib}
\bibliography{mybibliography}
\end{document}